\documentclass[runningheads]{llncs}
\usepackage{graphicx}
\usepackage{algorithm}
\usepackage{algpseudocode}
\usepackage{graphicx}
\usepackage{float}
\usepackage{amsmath}

\usepackage{xcolor}

\usepackage[colorlinks = true,
            linkcolor = black,
            urlcolor  = blue,
            citecolor = black,
            anchorcolor = black]{hyperref}

\newboolean{showcomments}
\setboolean{showcomments}{true}

\ifthenelse{\boolean{showcomments}} {
  \newcommand{\nb}[3] {
    {\color{#2}#1}
    {\color{#2}{#3}}
    }
  }
  {\newcommand{\nb}[3]{}}

\begin{document}
\title{Starkit: RoboCup Humanoid KidSize 2021 Worldwide Champion Team Paper}
\titlerunning{Champion Team Paper 2021}

%
\author{Egor Davydenko \and
Ivan Khokhlov\and
Vladimir Litvinenko \and
Ilya Ryakin \and
Ilya Osokin \and
Azer Babaev
}
\authorrunning{Team Starkit}
%
\institute{Team Starkit, Moscow Institute of Physics and Technology, Russia
\email{robocup.mipt@gmail.com}\\
}
\maketitle              
\begin{abstract}

This article is devoted to the features that were under development between \textit{RoboCup 2019 Sydney} and \textit{RoboCup 2021 Worldwide}. These features include vision-related matters, such as detection and localization, mechanical and algorithmic novelties. Since the competition was held virtually, the simulation-specific features are also considered in the article. We give an overview of the approaches that were tried out along with the analysis of their preconditions, perspectives and the evaluation of their performance.

\keywords{Robotics \and Simulation \and Computer Vision \and RoboCup}
\end{abstract}
\section{Introduction}

The main goal for these years was to improve the localization and motion stability. Also we managed to benefit from the stability by building more complex behaviour algorithms on top of it. Some of these features helped us to take first place in \textit{RoboCup 2021 KidSize Humanoid League}.

This year, the competition took place in simulator \textit{Webots}, which was a challenge for the competitors. We have devoted a lot of time and effort to create a simulation model of our robot that will be similar enough to the real-world one. Moreover, organizers have provided a very precise grass model. These factors allowed our robots to play in simulation almost with the same strategy and movements as in the real world. We believe that simulated games with automatic referee will intensify the strategy development for the competitors. Also, we are glad to announce the publication of our model\footnote{\href{https://drive.google.com/drive/folders/1Q6x5QzOxln3-fHeHYnt-hTPlJnfus896?usp=sharing}{Google Drive}} and docker image\footnote{\href{https://hub.docker.com/repository/docker/starkit/v-hsc2021}{Docker Hub}} to provide the ability to compete with our robots.

Speaking of the tournament, we have significantly improved our software during the competition. During the group stage we have had severe problems with the communication of \textit{Webots} and docker image, but with rapid and valuable support from the technical team we have managed to solve them. During the playoff, our robots have scored $32$ goals and received only $3$ and achieve the most effective game in our league by scoring $23$ goals in one game vs \textit{EDROM}. The most exiting game was the final one\footnote{\href{https://youtu.be/_9q26QjDluw}{YouTube Final Game}} with \textit{MRL-HSL} from Iran, which was a challenge. We suppose that changing strategy to more defensive and using opponent avoidance gave us a slight advantage. Thus, we were capable of winning, despite the fact that the \textit{MRL-HSL} robots walked faster and had a stronger kick.

We could highlight the following technical and algorithmic features behind the performance of our robots in the tournament.

\begin{itemize}
    \item Due to wide field of view of the cameras of our robots they were capable of fast ball detection. So it was possible for them to instantly start the movement towards the ball when it was necessary, while the opponents had to collect data about the surrounding for some time.
    \item The kick movement was fine-tuned to be robust and stable, leading to the predictable ball movement. So the pass game was possible, while the stronger kick of the opponents varied in the ball travelling distance.
    \item Fast localization of the robots made it possible to walk precisely to the necessary position for the while receiving the pass. Moreover, the robots were rapidly recovering after falling, meaning effectively no time to localize from scratch.
    \item Our robots were able to prevent collisions with teammates and opponents due to the stereo vision-based obstacle avoidance and precise localization. Also it was possible for them to follow the most active action as the game was unfolding.
\end{itemize}

Also, we work on developing robotics education \cite{edu} in our institute.

The paper consist of the following sections: robot model description is given, the wide-angle stereo vision and corresponding localization are introduced and lastly novel motion planning and falling prevention algorithms are presented.

\section{Model}
\begin{figure}
     \centering
     \begin{minipage}{0.45\textwidth}
         \centering
         \includegraphics[width=\textwidth]{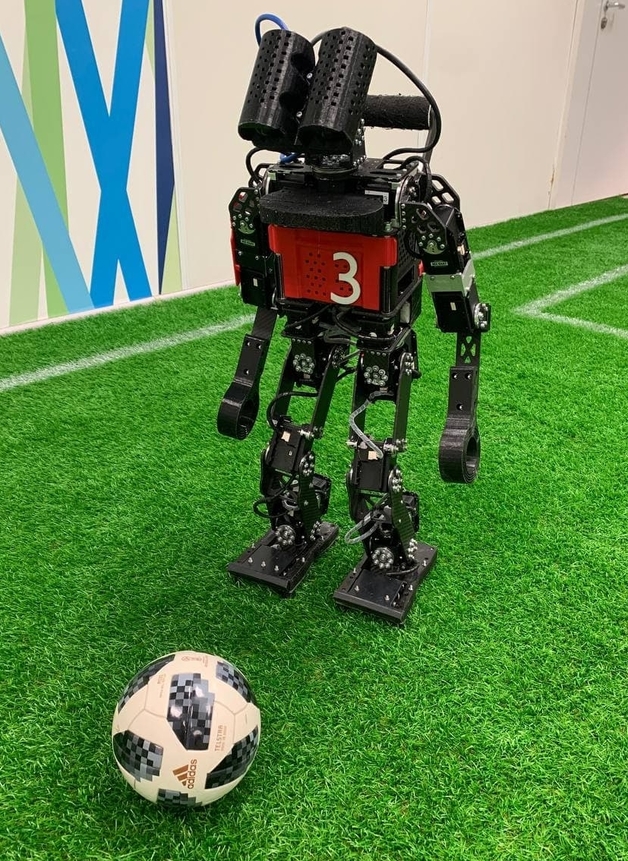}
         \caption{Real robot}
         \label{fig:model_real}
     \end{minipage}
     \hfill
     \begin{minipage}{0.45\textwidth}
         \centering
         \includegraphics[width=\textwidth]{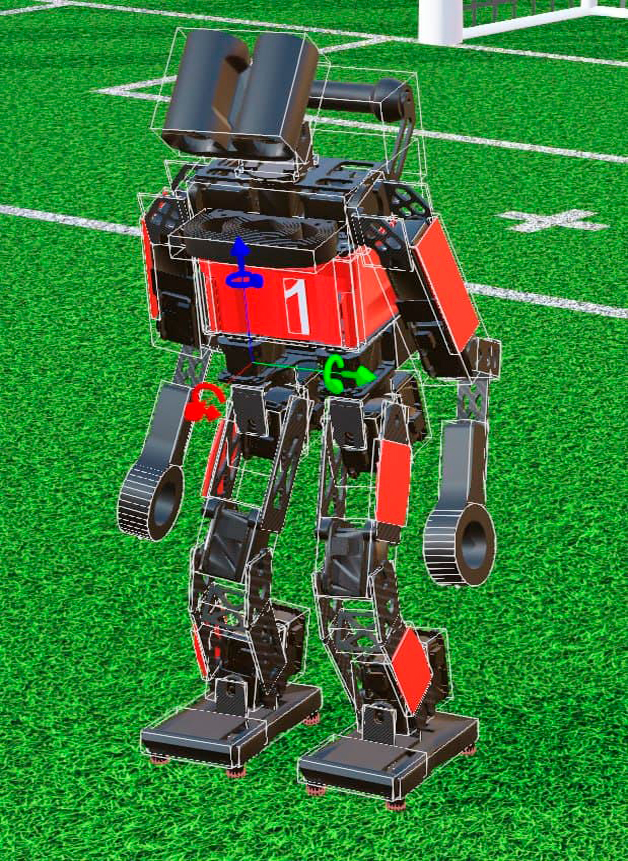}
         \caption{Robot model in \textit{Webots}}
         \label{fig:model_webots}
     \end{minipage}
\end{figure}
\subsection{Model and electronics changes}
The initial robot structure at the beginning of $2019$ was fully inherited from \textit{Rhoban team}, the world champion at that time. This model as a product of several years of development in the \textit{University of Bordeaux} had great robustness, and it gave us an invaluable amount of knowledge about building a good robot for KidSize league competitions.

Its main construction element is a flat milled aluminum plate. Real game tests demonstrated that this structure is prone to small irreversible deformations, often leading the damaged robot to look intact but performing worse.

Thus, we decided to switch to a milled carbon structure, which is both stiffer and lighter, redesigning some parts to better fit the carbon milling process. Also, we installed support bearings at each limb, including head, to mitigate the servo output shaft damage, often occurring when the falling robot hits objects with its head.

Also, we developed and installed a novel on-board power controller, allowing us to change the battery of the robot without restarting it. This controller relies on LTC4228 IC-based line switching, and we are happy to share it via our \textit{\href{https://github.com/StarkitRobots}{GitHub}}.

\subsection{Real2sim}
The robot model was exported from \textit{SolidWorks} to URDF\footnote{\url{http://wiki.ros.org/sw_urdf_exporter}} and then to PROTO format. We ensured that all the parameters, e.g. masses, inertia tensors, are close to the real values. This approach along with the use of the same material appearances where possible, helped us to get the model not only physically but also visually close to the real robot. 

\begin{figure}
    \centering
    \includegraphics[width=\textwidth]{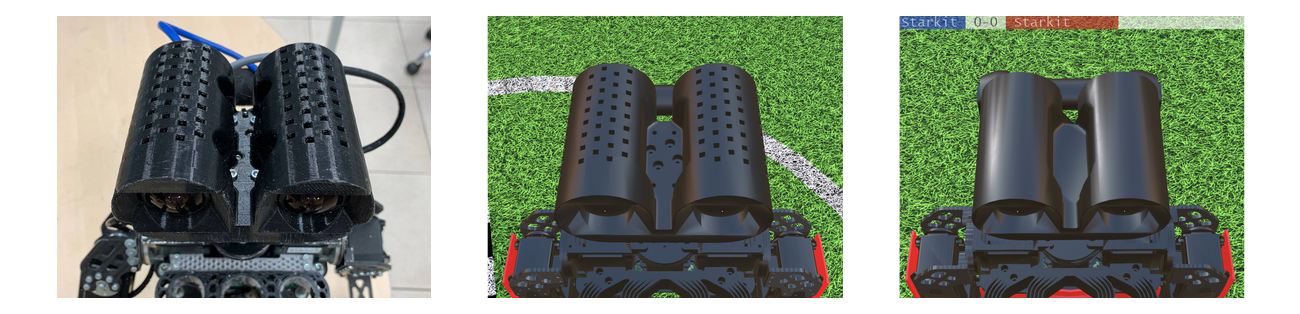}
    \label{fig:head}
    \caption{Head: from real to simplified}
\end{figure}
The most difficult for us was the simplification according to the Model Specifications
document\footnote{\url{https://cdn.robocup.org/hl/wp/2021/06/v-hsc_model_specification_v1.05.pdf}}. An example of the head simplification is given in \ref{fig:head}. Since there is no need for heat dissipation in the simulator, all the ventilation holes were removed. Wires, screws and their landing nests were also considered redundant for the simulation.

\begin{figure}
\centering
    \includegraphics[width=\textwidth]{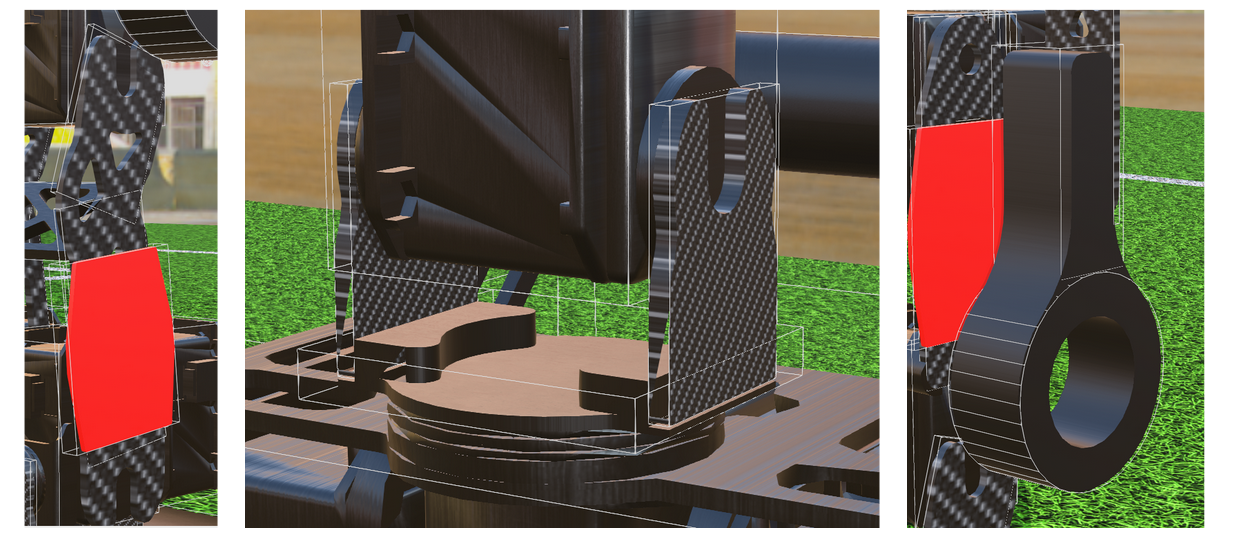}
    \label{fig:boundings}
    \caption{Bounding objects for legs (left), neck (middle) and arms (right)}
\end{figure}
Another problem was to fit the parts with complex shapes in the simple bounding objects. Lots of
time was spent to ensure that they are close enough to the original details, that there are not too many of them, and it is
possible to turn on the \textit{selfCollision} and do not ruin the motion because of the rough shapes \ref{fig:boundings}.
This robot model won the \textit{Best Humanoid Model Award}.

\section{Vision}
\subsection{Wide-angle lens}
Prior to 2020, our robot was equipped with a regular lens with approximately $45^{\circ}$ Field of View (FoV). This relatively narrow FoV forced the robot to constantly move the head with moderate velocity to scan the surroundings trying to find a ball or opponent robots, inherently emulating a large-FoV camera with constantly moving narrow-FoV camera.
This leads to the drawbacks, causing game quality degradation:
\begin{itemize}
    \item inability to rapidly detect the ball
    \item inability to perceive large structures on the field (prolonged goal lane / border lane / central circle)
\end{itemize}

At the end of 2019 we decided to switch to a wide-angle setup, and started to test it directly after the \textit{RoboCup Asia-Pacific 2019} event. We tested various lens setups and picked up the \textit{Beward BL0220M23} lens model, giving approximately $135^{\circ}$ FoV on our FLIR camera. This FoV allows the robot to see the ball in almost all required view angles using only the small panning motion of the camera, allowing the robot to react to the ball movements more rapidly.

But the wide-angle lens setup immediately leads to resolution and rectification problems. The first is the smaller pixel size of objects compared to a narrow FoV imaging.  For an “classic” narrow FoV imaging, we use a $720\times540$ resolution of a captured image with a $2\times2$ binning on a camera (which is capable of $1440\times1080$ raw image capturing). This binning allows us to lower the exposure time and to move a camera relatively fast with a little motion blur. At this binned resolution, the ball at the center of a robot’s sight has the pixel size of approximately $40\times40$ pixels. This size is enough to robustly detect the ball with our current convolutional neural network-based pipeline. But with $135$ degrees FoV at the same $720\times540$ input image resolution the pixel size of a ball is less than $20\times20$, which is below the robust detection threshold. To mitigate this problem, we switched to full $1440\times1080$ capturing with a lower camera panning speed. This resolution can still provide the reasonable pixel size of a ball in the center of a FoV of wide-angle lens, but with a cost of slightly higher motion blur and noise and larger exposure times.

\subsection{Vision pipeline parallelization and optimization}
Our “classic” vision pipeline prior to the end of 2019 was a slightly modified version of a highly successful vision pipeline of the \textit{Rhoban team}, used by them to win the \textit{RoboCup 2019} competition. This pipeline uses the $640\times480$ or $720\times540$ captured image size and is capable of processing it at 30 frames per second using a strictly “series” processing architecture. The series architecture means that each vision processing step can be performed only after the previous processing step is finished. For example, it’s possible to perform the goal post detection only after the ball detection step is done. According to Rhoban’s naming, each pipeline processing step is called a “filter”, and thus the “classic” vision pipeline is the series connection of different vision filters that runs strictly one after another connected by the corresponding inputs and outputs. Almost none of the vision filters we used at that time were capable of being parallelized, and ran mostly in single-core mode with some minor exceptions to truly OpenCV-based well-parallelized filters like color space conversion.

After switching from $720\times540$ to $1440\times1080$ capturing with a wide-angle setup, the throughput of this “series” vision pipeline degraded to less than $7$ frames per second, not allowing us to play a normal dynamic game. In order to optimize the vision pipeline performance, we rewrote it using a hybrid “series-parallel” architecture with an automatic filters parallelization. 

At the robots' software boot time, the vision pipeline governor loads its structure from a JSON file and automatically parses it to estimate which vision processing steps (filters) can be done in parallel (i.e. ball detection and goal posts detection). This step has a name of “filter’s batch”. The criterion of the possibility of the parallel processing for the group of vision filters is all their input dependencies that are satisfied simultaneously. We use the \textit{std::async} multi-threading to run all the filters in a batch in separate threads and wait for all the filters in a batch to finish. This pipeline allowed us to better utilize the multi-core architecture of an on-board PC, giving approximately 10-12 frames per second processing. 

To further optimize the vision pipeline, we divided processing steps that required a full-sized input image and filters that can work on downsampled input image with small robustness degradation. An example of such filter is the goal posts detector: they are significantly larger than the ball and can be detected well enough on a downsampled $720 \times 540$ image. In addition we parallelized all the possible filters with the OpenCV “parallel for” approach.

The second major drawback of a wide-angle lens setup we faced is the necessity to undistort the input image. The raw wide-angle image has a heavy radial distortion, causing straight lines to be bent significantly, especially at the borders of an image. At first sight, OpenCV library has a mature set of tools to perform the distortion correction, widely used by the robotics community. But the resultant undistorted image will be significantly larger than the captured image, causing the throughput of a vision pipeline to degrade twice: because of the undistortion time and because of the larger image size that will be processed in all the subsequent filters. For example, for the $135$ degrees FoV lens with a $1440\times1080$ capture, the rectified image will have a size bigger than $2880\times2160$ to cover all the pixels present in the captured image with the same resolution at the center of an undistorted image (assuming we are still using pinhole camera model with a flat imaging plane). 

To mitigate this, we modified the underlying math of the vision filters for most of them to perform directly on the raw wide-angle distorted image. For example, our ball detection routine performs its job on a distorted image, and only undistort the final position of several ball candidates, not the full image. So there is no classical undistortion filter at the beginning of our vision pipeline.

After all of these optimization steps, our wide-angle vision pipeline throughput is brought back to $20$ frames per second which (at least for us) is enough for a moderate dynamic gameplay, and only a small panning of a head is needed to detect the ball at all the required positions.

\subsection{Stereoscopic camera setup}
The classic approach to detect the opponents at the KidSize league is to use some CNN/DNN processing or semantic segmentation to find the robots on the captured image, and then estimate their field positions from monocular image position utilizing the fact that the opponent robot is standing on a flat ground. 

After some testing, we were not satisfied with this approach. Most of the robots in the league use some sort of lightened legs and thin feet, so only the body of a robot can be detected robustly enough but not the legs (at least with our detection approaches), especially when the robot is in front of the observer. This leads to the large ambiguity of a robot’s feet position estimation on a monocular image. But the ability to robustly estimate the opponent’s field position is mandatory for obstacle avoidance and ball kick direction calculation for better strategy.  Also, to detect the goal posts, goal net, legs of a referee and other obstacles the multi-class segmentation or detection DNN is needed with heavy processing time requirements, requiring to use some hardware neural networks accelerators like \textit{Intel Neural Compute Stick} or its analogues. 

As an attempt to fuse the monocular vision data and calculated geometrical data we have implemented a pipeline that adds the predicted depth as an additional channel to the TinyYOLO v3 NN detector, which resulted in a paper \cite{yolo}.

After that we decided to use a different approach and install a stereoscopic wide-angle vision system, being the first to do so in the KidSize league to our knowledge. 

This stereoscopic setup allowed us to robustly detect all the possible obstacles by their shape and size, not appearance or color, and estimate the distance to them with sufficient precision. This approach forced us to put large effort into solving camera synchronization issues and stereoscopic processing issues to keep the vision pipeline throughput fast enough.

Currently, we use a pair of \textit{FLIR BFS-PGE-16S2C} cameras with $62$ mm baseline. We use a hardware synchronization feature of these cameras, making one of it act as a master, generating the sync signal during the exposure time, and the other camera acts as a sync slave using the sync signal as a trigger input. Both frames are being augmented with a hardware timestamp. The vision pipeline input routine was modified to support two capture inputs, and the special synchronization governor was implemented to estimate the difference between timestamps of input frames and re-sync the capturing if one of the captured frames was missing due to high CPU load or another issues. Utilizing the fact that the FLIR camera has a global shutter feature, this setup allows us to get a good quality stereo pair even in the case of a fast-moving robot. 

The weight of a large head with a dual camera setup causes our robot to be a little less stable during walk and kick. We are mitigating this with the “active falling” and other walk stabilization techniques.

\subsection{Fast stereoscopic vision processing}

To keep the vision pipeline fast enough, we use downsampled $720 \times 540$ images as an input to the stereoscopic processing filter. The stereoscopic processing filter is an optimized custom-written routine doing all the stereoscopic processing steps in one filter. Currently, we use classic OpenCV rectification and disparity estimation, followed by custom disparity to point cloud processing combined with voxelization via binning and filtering. Then we use a ground plane estimation using RANSAC algorithm, and a combination of 2D connected components analysis and Point Cloud Library KD-Tree based EuclideanClusterExtraction \cite{RusuDoctoralDissertation} to detect obstacles as objects which are protruded by some threshold from a detected ground plane.

\begin{figure}
\centering
\includegraphics[width=340px]{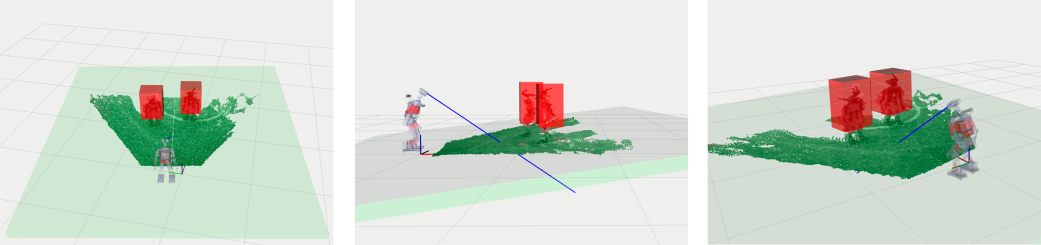}
\caption{Three different views of a real game example with two opponent robots being detected as obstacles in a 3D point cloud constructed from stereoscopic image using algorithms described above. Pictures are taken from a real-time point cloud feed by the robot, on-board processing speed is more than 10 FPS}
\label{fig:stereo_obstacles}
\end{figure}

Our on-board PC (currently the \textit{Intel NUC i7 gen9}) is capable of performing this stereo pair processing at 10-12 frames per second. As a comparison, classic ROS stereo processing, being integrated in our vision pipeline, gives about 2-3 FPS under the same conditions. So we abandoned it and used our less precise, but faster solution. Currently, we use ROS only for visualization of the resultant point cloud and obstacles for debugging purposes. (Sample of visualization is given on figure \ref{fig:stereo_obstacles})

To integrate these 10 frames per second stereoscopic processing in our project, we once again modified the vision pipeline to support filters that operate at the frequency that is a divider of the capturing frequency. Currently, the main pipeline is done at 20 FPS, and stereoscopic processing at 10 FPS. Our tests show that it is not necessary to detect the obstacles faster than 10 times per second because none of the current robots at the league move so fast that the detection does not work properly. But this holds true only in the case of a wide-angle stereoscopic vision setup that we have. If the robot needs to rapidly pan and tilt its head to cover all the required large field of view with a relatively narrow field of view stereoscopic camera, we assume that the stereo pair processing should be faster in order to not to miss important data.





\section{Localization}
\subsection{Line-based localization on narrow FoV}
The goal post detection-based localization that we used in 2019 (that were picked from Rhoban 2018 code release) was not robust enough even at the RoboCup 2019 competition. To find an alternative we switched to a field lane detection based approach. 

We implemented a custom vision filter for field lines detection. The main idea was to use some prolonged "global" features instead of small "local" ones to make the detection more robust. Our filter is capable of detecting two major types of features: a) the single field line and b) the corner of two field lines. 

Due to prolonged structure, field lines can be detected more reliably in the presence of noise and motion blur than the goal posts or other local features. Also in comparison to a goal posts or other point-like features with no distinct orientation information (we can observe a goal post from any angle as it will look the same), the lines and corners embed additional orientation information allowing the particle filter to converge faster due to less potential orientation ambiguity.

\begin{figure}
\centering
\includegraphics[width=300px]{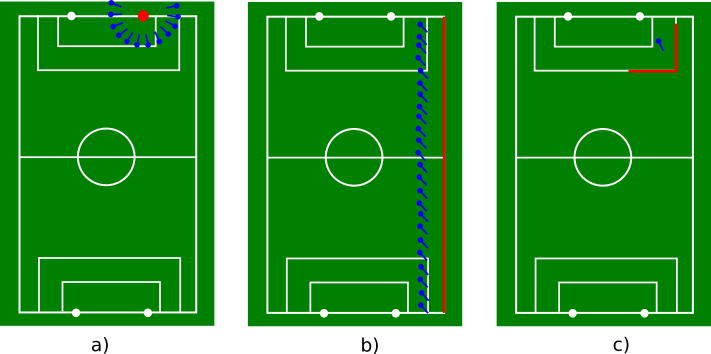} 
\caption{Possible positions of a robot (blue) to see the observation detected (red). Goal post (a): high orientation ambiguity / moderate position ambiguity. Line (b): very low orientation ambiguity / high position ambiguity. Corner (c): very low orientation ambiguity / very low position ambiguity. The corner is the most valuable observation among listed ones.}
\label{fig:localisation_potentials}
\end{figure}

At the first step, our filter tries to find any local features on the image that can be treated as the part of a straight line. Similar to the classic Sobel edge detection approach, we do two passes of a sliding window convolution - one in the horizontal, and one in the vertical direction. Our sliding window consists of three regions. The horizontal pass sliding window consists of three rectangles - left, middle and right. The middle rectangle corresponds to the line being detected, its size is picked up at each image location according to the expected line width at this field position. Left and right rectangles correspond to the regions to the left and right of a line being detected. The middle rectangle should be white and bright, and the left and right rectangles should be dark and green. The estimation of how this current image position fits this criterion is done via score function using integral image processing similar to ball detection sliding window routine. After this sliding window step, we get two images that we call a horizontal and vertical “heatmaps” for a horizontal and vertical pass separately. 

At the second step, we perform a Non-Maximum Suppression (NMS) \cite{1699659} algorithm to detect the local maximums of these heatmaps. This local maximums correspond to the middle pixels of a line being detected. To speed up the processing, the sliding window and NMS algorithms are performed not on each image row/column, but with decimation.

At the third step, these local maximums act as the inputs to the classic OpenCV Probabilistic Hough Transform \cite{Matas1998ProgressivePH} algorithm to detect the candidates of the line segments. Its result is then filtered to join the separate short segments into prolonged straight lines and to remove the duplicated lines if there are any. If the length of a line is above the threshold, it will be reported to the particle filter as a “line” observation.

At the fourth step, the resulting array of detected prolonged lines is checked for 90-degrees intersections. The basic type of intersection is the L-corner, and this is the type of observation that will be reported to the particle filter as a single “corner” observation. The T- and X-cross of field lines will be reported to the particle filter as two or four “corner” observations respectively.

As the result, each field line detected on the image is reported to particle filter twice: as an endless line observation, and as the part of a corner if this line intersected by another at the appropriate angle close to $90^{\circ}$. This gives a sufficient number of observations for the particle filter to quickly converge and keep track of the robot’s position, even in the narrow-angle FoV vision setup. The overall structure of our localization pipeline is given below:

\begin{figure}
\centering
\includegraphics[width=300px]{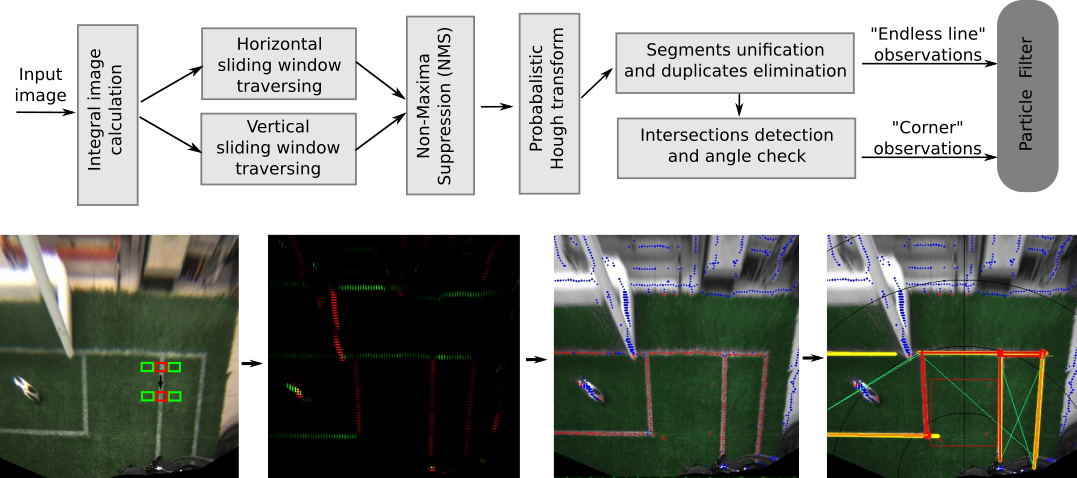}
\caption{Line/corners detection pipeline}
\label{fig:localization_structure}
\end{figure}

\subsection{Lane-based localization in wide-angle lens setup}
After switching to wide-angle lens setup, we extended our line-based localization approach to the large FoV imaging. Line detection algorithms require the lines on the image to be straight for them to be properly detected. This is not the case in the wide-angle camera setup, thus proper image rectification is needed. The straightforward wide-angle image rectification is too time-consuming and gives a large of the resultant image.

So for a wide-angle setup, we switched to the so-called “birdview” approach. Using camera position and orientation from IMU and forward kinematics, with assumption that the field is flat, we can calculate such a perspective transform that maps image pixels from the current robot’s point of view to a point of view of a new virtual camera, observing the field from the top and orthogonal to the field surface. This approach becomes especially efficient with the large field of view of a wide-angle camera setup, giving a good overview of the robot’s surrounding. After this transform, all non-planar objects (other robots, goal posts, etc.) are heavily deformed, losing their shape, but the field lines have constant width and proper geometry. 

Due to the relatively large, “global” scale of field lines compared to the small size of other "local" features like ball or goal posts, the field lines can be robustly detected on a birdview image using a relatively low-resolution image. Also lowering the resolution acts as an additional “low-pass” image filtering to remove noise and small field line discontinuities due to paint defects and other factors. Using the fact that the amount of pixels to be processed by undistortion routine depends on the undistorted image size and is almost independent from the input image size, a fast undistortion of a large input image to a low-resolution undistorted one could be performed, followed by a perspective transform on the same low resolution. We use a $640 \times 480$ output image for a birdview, and this processing can still be performed at more than $30$ frames per second on the board CPU. 

Using this technique, thanks to the large FoV of a wide-angle setup, the robot could localize itself robustly in approximately $2$ seconds after a complete localization loss.

\subsection{Modeling of a wide-angle lens in Webots simulator}
As RoboCup 2021 Worldwide took place in simulation, we had to model our wide-angle camera setup in a simulated environment. The default camera model that Webots 2021b provides can have a planar (rectilinear) and spherical projections, and theoretically speaking only the spherical projection is suitable to simulate the camera with a wide field of view. But at the time of the competition our test have shown that the projection mode of a simulated camera in "spherical mode" is actually a cylindrical projection, and this problem was not solved during the competition. So we were forced to emulate our wide-angle lens setup with a large field of view camera with a planar (rectilinear) projection followed by a simulated heavy radial distortion. This approach limits the actual field of view of a simulated camera and introduces some noise at the borders of the resulting image. Thus, our simulated camera image has a lower field of view than a real-world one (100 degrees instead of 135 degrees), and with a bigger inactive area (we use a procedural-calculated mask to cover this area with a solid black color to not to confuse our vision algorithms, see Fig.3). So the vision system in Webots simulator was modelled with worse characteristics than the real-world one. 

\begin{figure}
\centering
\includegraphics[width=340px]{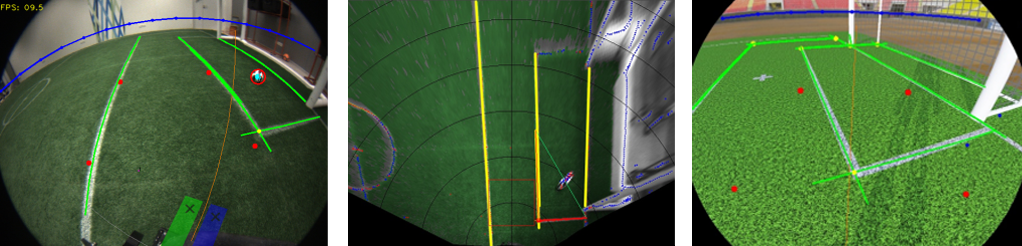}
\caption{\textbf{Left:} example of an input wide-angle image from a real robot with marked detected ball (red circle) and field lines (green). \textbf{Middle:} corresponding image after "birdview" transform with marked detected lines (yellow) and detected corner (orange/red/green triangle) constructed from NMS processing output (blue/red dots). \textbf{Right:} example of a simulated image from Webots simulator with a rectilinear projection followed by a radial distortion to mimic the real-world wide-angle image}
\label{fig:birdview}
\end{figure}

\section{Strategy}
\subsection{Ball path planning}
\begin{figure}
    \centering
    \includegraphics[width=0.49\textwidth]{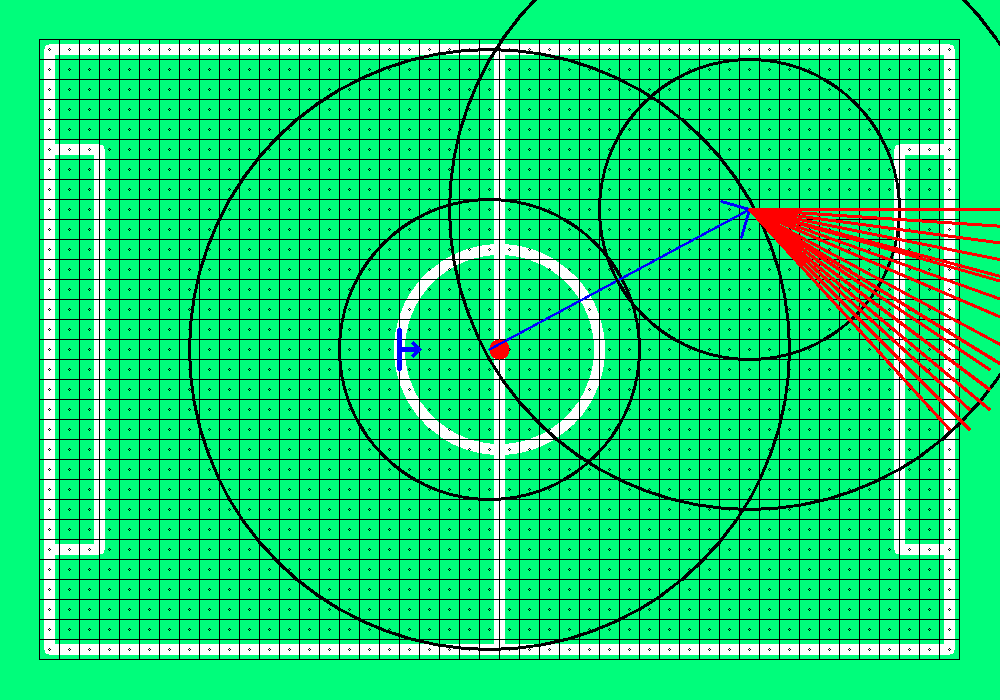}
    \includegraphics[width=0.49\textwidth]{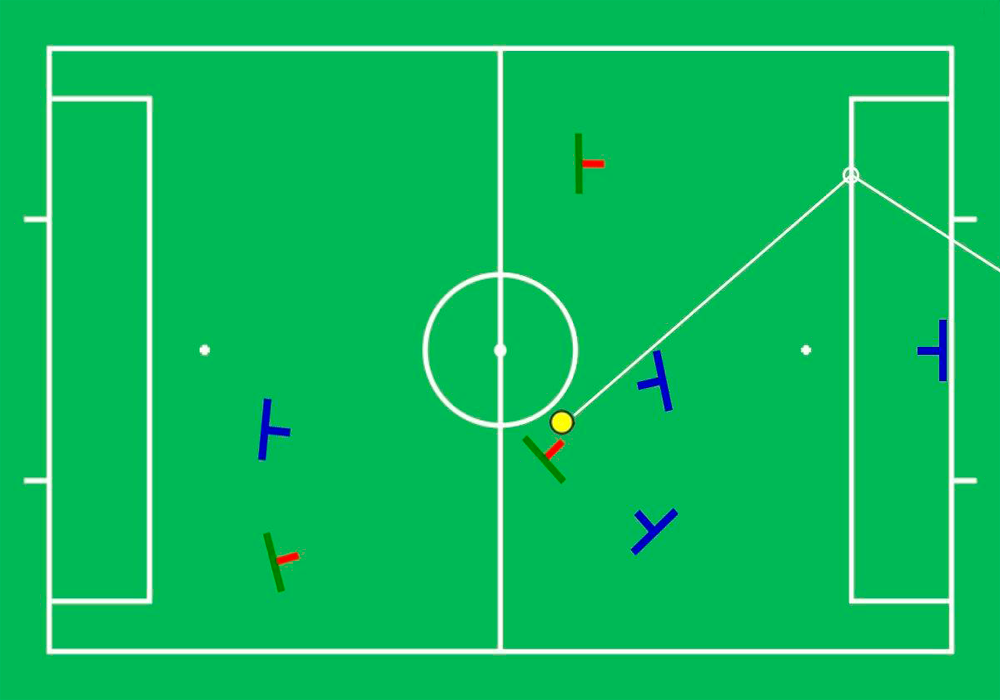}
    \label{fig:field}
    \caption{\textbf{Left:} Graph used for path planning. The centers of the grid cells define the vertices. Edges are defined implicitly by each pair of the vertices that (approximately) lie within the predefined kick distance from each other. \textbf{Right:} Example of the algorithm output.}
\end{figure}

Improvements in localization and vision systems allow us to construct more complex and smart strategy. We decided to work on the approach to choosing kick direction. The algorithm that we used is described in \cite{ap}. We introduce a  graph~\figurename~\ref{fig:field} by discretizing the field. Each cell is a square of $10\times10$ cm, so  there are $90 \times 60 = 5400$ cells overall. Then, given the ball position, the cell with a center that is the closest to the ball position is identified, and it is considered as the first vertex of the graph. Then the edges of the graph are generated with the length of the possible kicks and the shortest path is calculated with the $A^*$ \cite{hart1968formal} algorithm with the cost (Alg.~\ref{alg:costFunc}) and heuristic functions (Alg.~\ref{alg:hFunc}). 
With this algorithm the robot prefers to play a pass and to move ball to the free zones.

\begin{algorithm}
\caption{Cost function}\label{alg:costFunc}
\begin{algorithmic}
\Function{computeCost}{robotPos, opponentsPositions, ballFromPos, ballToPos, firstKick}
\State $ballTravelTime\gets getLength(toPos, fromPos)/ballSpeed$
\If {firstKick}
    \State $timeToReachBall \gets calcTimeToApproachBall(fromPos, robotPos)$
    \If {intersectOpponent(fromPos, toPos, opponentsField)}
        \State \Return $timeToReachBall + ballTravelTime * 2$
    \Else
        \State \Return $timeToReachBall + ballTravelTime $
    \EndIf
\Else
    \State \Return $ballTravelTime$
\EndIf
\EndFunction
\end{algorithmic}
\end{algorithm}

\begin{algorithm}
\caption{Heuristic function}\label{alg:hFunc}
\begin{algorithmic}
\Function{hFunc}{teamMatesField, toPos, firstKick}
\State $timeToReachGoal\gets distToGoal(toPos)/ballSpeed$
\If {firstKick}
    \State $timeToApproachBall \gets calcTimeToApproachBall(toPos, teamMatesField)$
    \State \Return $timeToApproachBall + timeToReachGoal$
\Else
    \State \Return $timeToReachGoal$
\EndIf
\EndFunction
\end{algorithmic}
\end{algorithm}

We have used this algorithm for some games, but for now more straightforward algorithms like kicking directly to the goal are applied. The main problem is that the planer is sensitive to the noise in the ball position. We believe that at some point we will be able to improve the algorithm for it to givehttps://www.office.com/?auth=2 satisfactory results.

\section{Motions}

\subsection{Active falling}
We have noticed that the robots spend much time recovering from the falls, specially when standing up from the back. Thus, we introduced so called active falling, that is activated when the robot CoM leaves the support polygon. During this motion the robot throws out its arms in the direction of the fall\footnote{\url{https://youtu.be/ZZ7lhSuwT2o}}. After this, the robot start the stand up motion directly from this pose. For back stand up the recovery time decreased by the factor of $1.6$ and for the front one by the factor of $1.4$. This motion is harsh to the shoulder pitch servos, so we have developed flexible arms and added support bearings.

However, this motion is still too unstable and potentially harmful, so we are planning to improve it with better falling detection and processing.
\begin{figure}
    \centering
    \includegraphics[width=0.7\textwidth]{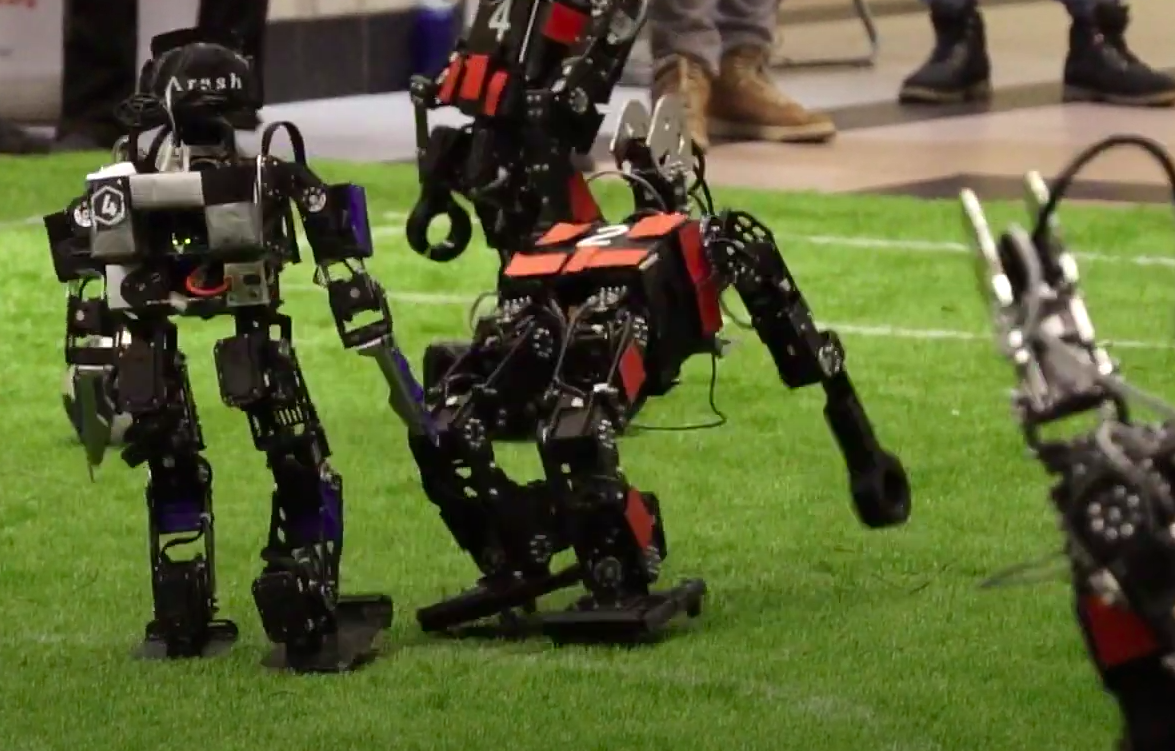}
    \label{fig:active_falling}
    \caption{}
\end{figure}

%
%
%
\bibliographystyle{splncs04}
\bibliography{biblio.bib}

\end{document}